\documentclass[letterpaper, 10 pt, conference, pdftex]{ieeeconf}  

\IEEEoverridecommandlockouts                              

\usepackage{graphics} 

\title{\LARGE \bf
SCU-Hand: Soft Conical Universal Robotic Hand for \\ Scooping Granular Media from Containers of Various Sizes
}

\author{Tomoya Takahashi$^{1}$, Cristian C. Beltran-Hernandez$^{1}$, Yuki Kuroda$^{1}$, \\ Kazutoshi Tanaka$^{1}$, Masashi Hamaya$^{1}$, and Yoshitaka Ushiku$^{1}$
\thanks{*This work was supported by the JST-Mirai Program Grant Number JPMJMI21G2, Japan.}
\thanks{$^{1}$OMRON SINIC X Corporation, Tokyo, Japan. 
        {\tt\small tomoya.takahashi@sinicx.com}}%
}

\usepackage[dvipdfmx]{graphicx}
 \usepackage{multirow}
 \usepackage[table,xcdraw]{xcolor}
 \usepackage{textcomp, gensymb}
\usepackage{amsmath}

\usepackage{color}
\usepackage{colortbl}

\usepackage[absolute,overlay]{textpos}

\begin{document}

\maketitle
\thispagestyle{empty}
\pagestyle{empty}

\begin{textblock*}{18cm}(1.5cm,0.5cm) 
    \tiny 2025 IEEE International Conference on Robotics and Automation (ICRA2025). Preprint. Accepted January 2025. \textcopyright 2025 IEEE.  Personal use of this material is permitted.  Permission from IEEE must be obtained for all other uses, in any current or future media, including reprinting/republishing this material for advertising or promotional purposes, creating new collective works, for resale or redistribution to servers or lists, or reuse of any copyrighted component of this work in other works.
\end{textblock*}

\begin{abstract}

Automating small-scale experiments in materials science presents challenges due to the heterogeneous nature of experimental setups. This study introduces the SCU-Hand (Soft Conical Universal Robot Hand), a novel end-effector designed to automate the task of scooping powdered samples from various container sizes using a robotic arm. The SCU-Hand employs a flexible, conical structure that adapts to different container geometries through deformation, maintaining consistent contact without complex force sensing or machine learning-based control methods. Its reconfigurable mechanism allows for size adjustment, enabling efficient scooping from diverse container types. By combining soft robotics principles with a sheet-morphing design, our end-effector achieves high flexibility while retaining the necessary stiffness for effective powder manipulation. We detail the design principles, fabrication process, and experimental validation of the SCU-Hand. 
Experimental validation showed that the scooping capacity is about 20\% higher than that of a commercial tool, with a scooping performance of more than 95\% for containers of sizes between 67~mm to 110~mm.
This research contributes to laboratory automation by offering a cost-effective, easily implementable solution for automating tasks such as materials synthesis and characterization processes.

\end{abstract}

\section{Introduction}\label{sec_Introduction}
Laboratory automation (LA) through robotic systems has the potential to significantly improve research efficiency as it can attain higher throughput and precision by performing tasks faster and with greater consistency than manual methods~\cite{duros2017human}. 
In many disciplines, such as materials science, particularly solid-state materials discovery, it is crucial to automate small-scale experiments, dealing with small quantities and high variability, for processes such as synthesis and subsequent characterization of diverse materials. While the mechanization of several procedures has advanced~\cite{tom2024self}, material handling between steps is still often performed manually~\cite{chen2024navigating}. 

LA for these small-scale experiments presents unique challenges due to the heterogeneous nature of experimental setups. Containers and tools employed in these processes vary depending on specific experimental protocols and objectives. Therefore, robots must demonstrate the ability to adapt to these diverse conditions~\cite{burger2020mobile}. 
Furthermore, handling powders is challenging due to their unique dynamics. Currently, most automated solutions for handling powder are task-specific, such as dedicated dispensers~\cite{jiang2023autonomous}, robotic arms for powder grinding~\cite{Nakajima2022grinding}, powder weighing~\cite{kadokawa2023learning}, and even a fully automated system to handle several experimental procedures by creating an environment fully customized for robot arms~\cite{szymanski2023autonomous}. However, these approaches are limited to a fixed set of task conditions and can not easily adapt to novel conditions, such as changes in the container size.

   \begin{figure}[t]
      \centering
     \includegraphics[width=0.95\linewidth]{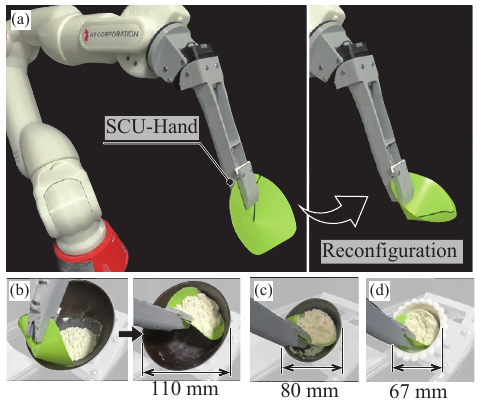}
      \caption{SCU-Hand: (a) A reconfigurable robotic hand that can fit containers of different sizes.  (b) scooping by maintaining full contact with the container surface via flexibility. (c) and (d) scooping from small containers. *In these figures, a colored sheet was used for the end-effectors, which was different from the sheet used in the experiment.}
      \label{fig_concept}
   \end{figure}
   
This study aims to automate the task of scooping powdered samples from containers of various sizes using a general-purpose robotic arm. This process involves the transfer of small quantities of synthesized material from spherical receptacles, such as mortars, to secondary containers for subsequent analytical procedures. Considering the small amounts of powder involved, it is crucial for the robot arm to scoop nearly all of the powder in a single action to significantly improve the efficiency of this task. 
This task is challenging because the robot's tool, i.e., the end-effector, must be precisely moved along the container's surface while maintaining consistent contact because fine powder escapes through the gaps between the end-effector and the container. Prior approaches to this type of task required the integration of force and torque sensors, visual recognition systems, and sophisticated control algorithms, including reinforcement learning techniques~\cite{pizzuto2022accelerating}. 
Furthermore, the variability of container dimensions requires either specialized tools for each container or a sufficiently small tool to accommodate all container geometries. 




To tackle scooping powder sample tasks, an end-effector must possess certain characteristics: It should be thin and rigid enough to easily slide under the powder while maintaining its shape and have a spoon-like form to prevent the powder from spilling. Taking that into account and inspired by the adaptive deformation of soft grippers~\cite{brown2010universal,shintake2018soft}, which increases contact area through contact, our approach includes flexibility as a design principle. Additionally, we focused on designing a conical surface that could be deformed from a circular sheet while preserving both flexibility and thinness. Our design meets these characteristics by using a single deformable sheet and a single actuator.

Our main contribution is a flexible end-effector called SCU-Hand (Soft Conical Universal Robot Hand). Its flexibility allows it to maintain contact with the container through adaptive deformation without the need for force sensing or reinforcement learning. 
Furthermore, the SCU-Hand can adjust its size according to the container via a reconfigurable mechanism, enabling scooping motions that work for various container types, as shown in Fig.~\ref{fig_concept}.
A thin, flexible sheet enables universal scooping of powders, viscous materials, and fragile objects, extending Universal Gripper~\cite{brown2010universal}' adaptability to amorphous materials.
Overall, this paper aims to automate real-world material synthesis, focusing on low cost and minimal installation time. It primarily develops and validates an end-effector designed to maximize the amount of powder scooped from various containers.

\section{Related Work}\label{sec_related_work}
In this section, we compare existing studies related to automating scooping tasks to establish the open challenges and requirements necessary to achieve this task.

\subsection{Manipulation of granular media}\label{sec_rw_scoop_task}


Prior research work focused specifically on scooping tasks includes:
Grannen et al.~\cite{grannen2022learning} propose a method for scooping a fragile object by combining a human spoon and two robot arms equipped with special end-effectors.
Tai et al.~\cite{tai2023scone} show a method for scooping granular media from a container using a hard ladle as a tool and a combination of vision and machine learning-based control. 
Bo et al.~\cite{bodesign} showed that scooping could be easily achieved while maintaining contact with the bottom of the container by connecting rigid plate-like spatulas with a compliant joint and actuating them by a tendon. 
However, none of these existing studies on scooping have addressed the task of scooping a small amount of powder from variously sized containers in a single attempt. While these studies may have the potential to scoop larger quantities through repeated actions, they rely on rigid end-effectors, making it difficult to maintain complete contact with the container, a requirement for efficient one-shot scooping. 


\subsection{Morphing mechanism of flexible sheet material}\label{sec_rw_soft_deform_mech}
Active deformation of a flexible material can be used to reconfigure the end-effector to fit the geometry of a given container.
The following is a comparison of existing deformation methods for thin plate structures.
Some mechanisms have been proposed to deform structures made of silicone or flexible plastic sheets by inflating them with air~\cite{niiyama2015pouch}. Ou et al.~\cite{Ou2016aeromorph} have proposed a method to deform them in various directions under pneumatic pressure by welding thin sheets of material in a specific pattern. 
Origami structure can transform a sheet material with a specific fold pattern from a flat state to a three-dimensional state~\cite{mitani2009design}\cite{li2019vacuum}. 
Keely et al.~\cite{keely2024kiri} show that a shape-changing kirigami-inspired end-effector increases teleoperation scooping efficiency: the three-dimensional deformation allows both scooping and pinching and stable object manipulation. 
Electroactive polymers~\cite{kofod2007energy,kubo2020simultaneous} and piezoelectric devices\cite{wang1999electromechanical}  can deform various shapes of the sheet, maintaining their thinness. 
Several mechanisms have been proposed to bend metal sheet structure by displacement differences ~\cite{yamada2017loop,takahashi2021two,xiong2023rapid}.

However, even if the end effector is reconfigured its size using these methods, it causes some problems in the scoop task. A concave shape similar to a spoon is desirable to transport powder stably. Methods such as~\cite{keely2024kiri},~\cite{takahashi2021two}, and~\cite{Ou2016aeromorph} achieve this, but ~\cite{keely2024kiri},~\cite{takahashi2021two} have large gaps between sheets, causing the powder to spill. Additionally, in~\cite{Ou2016aeromorph}, the thickness of the sheet is increased due to inflation by air, making it difficult to use for scooping tasks. On the other hand, 
~\cite{kofod2007energy,wang1999electromechanical,kubo2020simultaneous} bends a uniform surface, but the flexibility of polymer allows it to buckle due to friction when the end-effector slides on the container. The piezo-electronic actuator cannot achieve a sufficient large curvature, making it unsuitable for size reconfiguration. 
Therefore, an optimal solution would require a mechanism that balances maintaining a concave shape with the ability to deform its size efficiently without gaps or excessive thickness.
   
\section{Problem Formulation}\label{sec_problem}

\begin{figure}[t]
    \centering
    \includegraphics[width=0.85\linewidth]{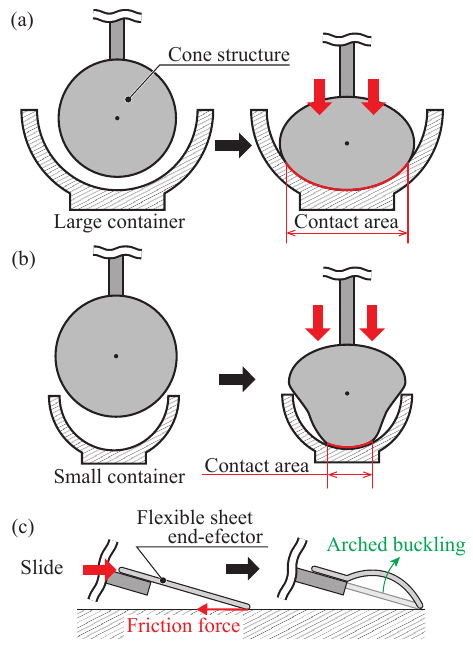}
    \caption{
    Different adaptive deformations caused by differences in end-effector size relative to the container size:
    (a) If the end-effector is smaller than the container, the contact area can be widened by contact with the container. (b) If the end-effector is larger than the container, insertion is possible, but the contact area becomes smaller. (c) Buckling of flat sheet structure with flexible material while sliding according to surface.}
    \label{fig_deform_in_container}
\end{figure}

\subsection{Task setting and assumptions}\label{sec_target_task}
Here, we defined the tasks addressed in this paper, which are primarily designed for use in LA and have the potential for broader applications.

\subsubsection{Robotic system and workspace}
We consider the robotic scooping system to be a general-purpose robotic arm with a single type of end-effector. The container is fixed to the ground. The robot arm has a different predefined motion for each container, though precise force and contact control are not employed.

\subsubsection{Container and powder types}
The system accommodates several types of spherical containers similar to mortars commonly used in laboratories, both larger and smaller than the end-effector.
Flour, a fine granular substance, was the primary material used. Additionally, materials with larger particle sizes, such as coffee powder and rice, were utilized. The amount of material in the experiment was sufficient for a single scooping operation.

\subsubsection{Scooping ability}
Our goal is to achieve the highest scooping amount possible in a single operation. 

\subsection{Mechanism requirement}\label{sec_mech_requirement}
In the following, we will state the requirements for the flexible end-effector necessary to achieve the target task.
\subsubsection{Thinness}
The end-effector must have a thin sheet-like structure to slide easily under the material. If it is too thick, it may push the powder out of the container, leading to spillage. 
\subsubsection{Size reconfiguration}
The end-effector's size needs to be reconfigurable to accommodate small and large containers, such as a controllable deformation (Fig.~\ref{fig_deform_in_container}).
\subsubsection{Prevent spilling of powder}
To securely hold the powder and prevent spillage, a concave shape without gaps is ideal. This ensures that the powder remains contained within the structure, minimizing the risk of spillage during handling.
\subsubsection{Anisotropic stiffness}
We aim to increase the contact surface with the container by flexibly deforming the end-effector when pressed against it, as shown in Fig.~\ref{fig_deform_in_container}~(a). Simultaneously, it is crucial to maintain sufficient stiffness to prevent buckling under friction or reaction forces from the material when the end-effector slides along the container surface, as shown in Fig.~\ref{fig_deform_in_container}~(c).

\section{Design Concept and Prototyping of the SCU-Hand}  \label{sec_design}

\begin{figure}[t]
    \centering
    \includegraphics[width=\linewidth]{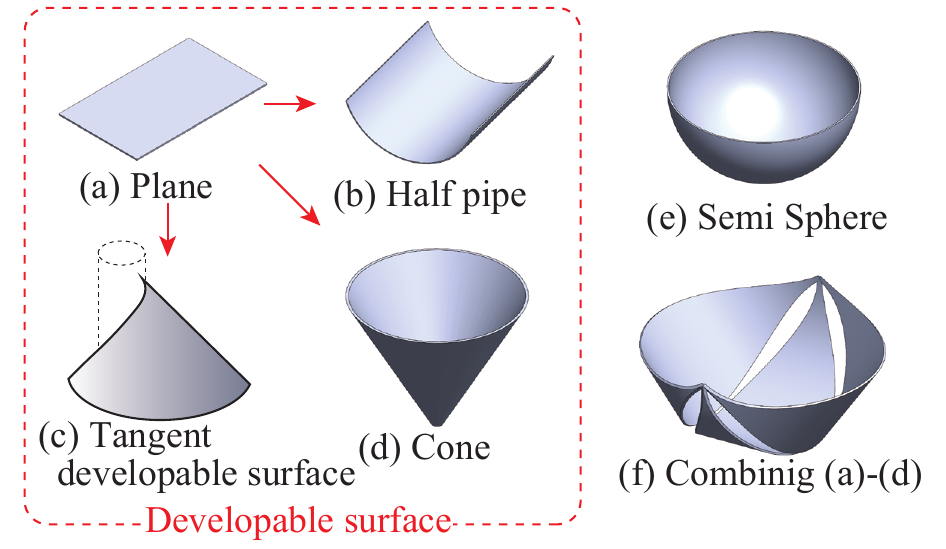}
    \caption{(a)-(d) are developable surfaces and can be deformed from a single plate. (e) requires a large elongation in the material}
    \label{fig_developablesurface}
\end{figure}

\subsection{Optimal sheet structure} 
We considered that a mechanism in which a flexible sheet structure deforms a three-dimensional surface could achieve requirements 1), 2), and 3). We explored the optimal shape by classifying mathematically defined shapes that can be deformed from a sheet. Such structures are mathematically defined as developable surfaces~\cite{lawrence2011developable}, which are shapes that can be deformed from a single sheet without stretching or tearing. There are only five types of surface that can be deformed from a continuous sheet, as shown in Fig.~\ref{fig_developablesurface} (a)-(d) and their combination (f). It is important to note that a semi-sphere shape like (e) is not a developable surface, which is why the kirigami structure~\cite{keely2024kiri} requires gaps in its sphere-like shape.
   
Among the available options, the concave shapes (c) and (d) are best suited for minimizing powder spillage. We chose the cone shape because it allows continuous adjustment of the bottom circle's diameter, making it ideal for variable size adaptation. While a concave shape can be constructed using a combination of multiple materials, there is a risk of powder residue accumulating in overlapping areas or joints. Therefore, the cone shape is considered more desirable, as it avoids such issues by maintaining a seamless structure.

\subsection{Basic principle of cone structure} 

\begin{figure}[t]
    \centering
    \includegraphics[width=\linewidth]{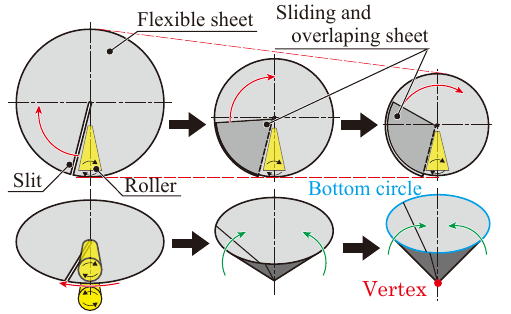}
    \caption{Size reconfiguration of a cone: initially slides one end of the circular sheet and morphs it into a conical surface by overlapping it, thereby reducing its diameter.}
    \label{fig_warping_of_cone}
\end{figure}

\begin{figure}[t]
    \centering
    \includegraphics[width=0.9\linewidth]{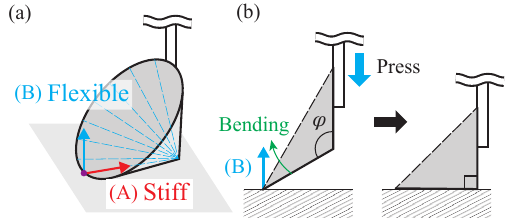}
    \caption{(a) Force generated on the cone structure during the scooping operation, (b) Contact tolerance. }
    \label{fig_stiffness_of_cone}
\end{figure}

A single flexible sheet structure that morphs into a conical shape (Fig.~\ref{fig_warping_of_cone}) can satisfy all the requirements in Sec.~\ref{sec_mech_requirement}: 1) it consists only of a single thin sheet, 2) its size can be reconfigured (bottom circle) by sliding and overlapping the circular sheet, 3) Then it morphs into a cone, a 3D shape, which keeps the powder contained towards the vertex, 4) The cone structure has anisotropic stiffness which is preferable for scooping task~(Fig.~\ref{fig_stiffness_of_cone}). Here, we explain the derivation of the deformation formula for controlling the mechanism's transformation and discuss the stiffness characteristics of the cone structure.

\begin{figure}[t]
    \centering
    \includegraphics[width=0.7\linewidth]{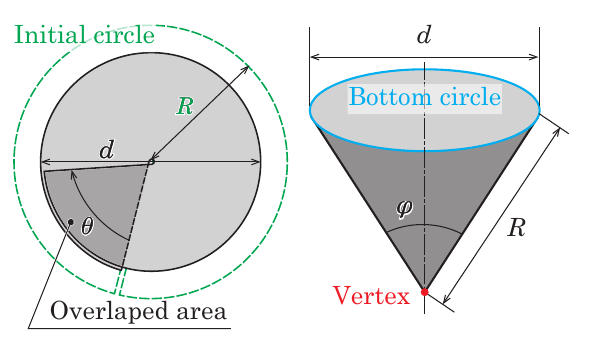}
    \caption{Schematic of relation between sliding angle $\theta$ and diameter $d$, and vertex angle $\phi$.}
    \label{fig_cone_parameter}
\end{figure}

The relationship between the slide angle $\theta$ and the dimensions of the conical structure is illustrated in Fig.~\ref{fig_cone_parameter} and expressed as follows:
First, suppose that initially, in a circular sheet with radius $R$, one end of the sheet is rotated by an angle $\theta$. This deforms the bottom circle into a conical shape with a $d$ diameter and an angle of $\phi$ at the vertex. In this case, $d$ decreases linearly with $\theta$ and is expressed by the following equation.
\begin{equation}
    d=2R(1-\frac{\theta}{2\pi})\label{eq_diameter}
\end{equation}
Since the length of $R$ does not change as $d$ is decreased in the conical structure, the theoretical minimum size of the end-effector is $R$. Therefore, the conditions under which an end-effector can be inserted for a container of diameter $D$ is
\begin{equation}
    d < D \quad (\text{where} \ R < D)\label{eq_condition_of_insert}
\end{equation}
From Eqs.~\ref{eq_diameter}, and~\ref{eq_condition_of_insert}, when inserting into a container of diameter $D$, theta should be set as follows
\begin{equation}
    \theta > 2\pi(1-\frac{D}{2R})\label{eq_theta_insert}
\end{equation}
Also, the vertex angle $\phi$ is determined by the ratio of $d$ to the initial radius $R$, 
\begin{equation}
   \phi=\arcsin(\frac{d}{2R})\label{eq_angleofvertex}
\end{equation}
Table~\ref{tb_Parameter_of_cone} shows combinations of $\theta$ and $\phi$ settings for reducing the initial diameter of 100 mm to $d$~=~90~mm, 80~mm, and 70.7~mm. Since the minimum value of $\phi$ is set to 90\degree, the minimum value of $d$ is 70.7~mm.

\begin{table}[t]
    \centering
    \caption{Example of a combination of dimensions of cone}
    \label{tb_Parameter_of_cone}
    \begin{tabular}{l|llll}
        \textbf{Diameter of bottom circle $d$ [mm]}     & 100 & 90  & 80  & 70.7\\\hline
        \textbf{Sliding angle $\theta$ [\degree]}       & 0   & 36  & 72  & 105 \\
        \textbf{Angle of vertex $\phi$ [\degree]}        & 180 & 128 & 106 & 90   \\ 
    \end{tabular}
\end{table}
   
   
The conical structure can satisfy anisotropic stiffness in Requirement 4). This stiffness arises because a sheet structure cannot simultaneously exhibit curvature in two orthogonal directions. 
For example, the half-pipe shape shown in Fig.~\ref{fig_developablesurface}(b) is resistant to deformation in the direction orthogonal to the arch-shaped bent cross-section. 
Similarly, a cone, having curvature around its central axis, demonstrates high stiffness in direction~(A) in Fig.~\ref{fig_stiffness_of_cone}(a). On the other hand, in direction~(B), deformation easily occurs because it is just a change in the initial curvature. 
This demonstrates that when the diameter of the bottom circle is smaller than that of the container, the contact to the container deforms the bottom circle into an elliptical shape, thereby increasing the contact surface and eliminating the need for whole 3D spherical surface recognition and shape control~(Fig.~\ref{fig_deform_in_container}).

\subsection{Design and fabrication of the SCU-Hand}

\begin{figure}[t]
    \centering
    \includegraphics[width=\linewidth]{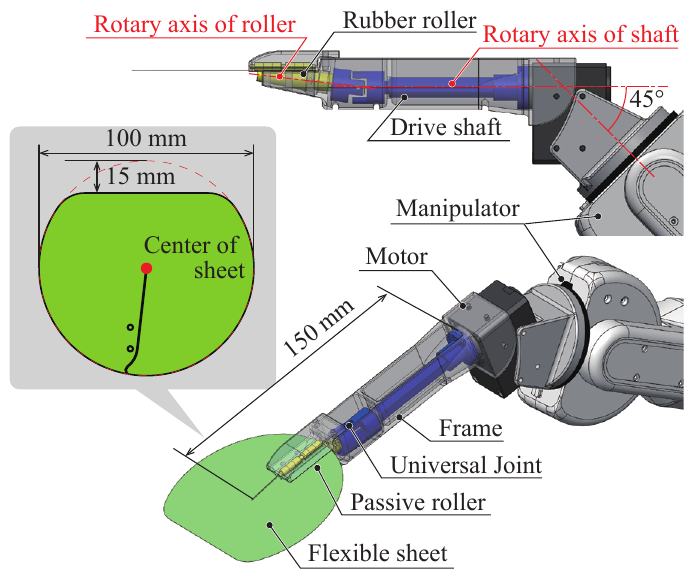}
    \caption{Proposed SCU-hand's configuration and rotation axis arrangement.}
    \label{fig_prototype_design}
\end{figure}

Fig.~\ref{fig_prototype_design} shows a prototype of the proposed end-effector to perform experiments on scooping tasks. The SCU-Hand comprises a flexible sheet, a frame, rollers, a drive shaft, and a motor. The prototype is mounted at a 45\degree angle relative to the manipulator. This prototype generates tilting of the sheet driven by the rotation of rubber rollers.
  
The shape of the flexible sheet is almost circular, but a flat edge is incorporated at the front to accommodate containers with flat walls (Fig.~\ref{fig_prototype_design}). Additionally, the end-effector is mounted 150~mm away from the motor to allow its insertion into a deep container. The rollers consist of a combination of a nitrile rubber roller and a free roller made of polyacetal, and the sheet is driven by pinching it with the two rollers~(Fig.~\ref{fig_warping_of_cone}). Since the amount of displacement depends on the distance from the center of the flexible sheet, a tapered rubber roller is used and connected to the drive shaft with a universal joint to tilt the rubber roller's rotation axis. The drive shaft is made of aluminum alloy, and the frame is 3D printed. The motor is a Dynamixel XM430-W350-R (ROBOTIS Co., Ltd., 2024), which is connected to the drive shaft by a flange also made by 3D printing. The total weight of SCU-hand is 138g.

   
\section{Experiments}
\label{sec_experiment}

\begin{figure}[t]
    \centering
    \includegraphics[width=\linewidth]{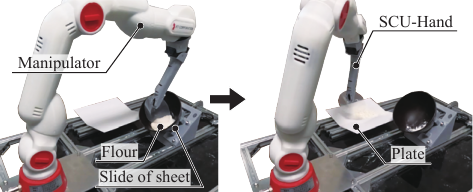}
    \caption{Experimental setup for scooping tasks.}
    \label{fig_experimental_setup}
\end{figure}

This study addresses the following questions: 1) Does adjusting the size of the end-effector according to the container size improve scooping performance? 2) Is the flexibility of the end-effector more effective compared to traditional rigid tools? 3) Does the performance vary depending on the type of scooped material? 
To answer the above questions, Experiment 1 involved performing scooping tests with four different containers while varying the size of the end-effector. To address questions 2 and 3, comparisons were made using a specific container to evaluate the effectiveness of the flexible end-effector against a rigid tool, and assess the proposed SCU-hand's performance when using different materials for the flexible sheet.

\subsection{Experimental setup}
First, the experimental setup, which is common for all the experiments, is described. The experimental environment is shown in Fig.~\ref{fig_experimental_setup}. A CRANE-X7 robot arm (RT Corporation, 2024) fixed to a frame was used. Our proposed end-effector, described in Section~\ref{sec_design}, was attached to the robot arm.
The proposed end-effector was designed to have a maximum diameter of 100~mm so that it could scoop from the 80~mm diameter of the container used in the previous study~\cite{Nakajima2022grinding}. For evaluation, the following materials were used: several spherical containers with diameters ranging from 67~mm to 110~mm were used. The inner surfaces of the containers are non-uniform in curvature, as they are commercial products.
The containers were mounted at a 45\degree tilt by a 3D printed base to ensure an effective scooping trajectory within the manipulator's range of motion.
Three end-effectors were evaluated. First is the SCU-Hand prototype, described in Section~\ref{sec_design}, with a polypropylene (PP) plastic sheet. Second, the SCU-Hand prototype but with a metal sheet (SUS304, t = 0.1~mm). Finally, a commercially available ladle (silicon, about 70~mm of width) was used in place of the SCU-Hand. Flour, coffee powder (medium-fine grind), and rice were used as target granular media.

In the experiment, the end-effector is inserted into the container while in contact with its inner surface, scoops up the powder, and then drops the scooped powder onto a plate next to it. The arm trajectory is set to follow multiple pre-defined waypoints, among which only those for scooping are fixed in a plane passing through the center of the container. After placing the specified amount of powder in the container, the container is vibrated to flatten the powder. The scoped amount was considered as the amount of powder dropped onto the plate. For each test, we recorded the average scoped amount of 10 trials.

The differences in conditions for each experiment are shown below.
\subsubsection{Verification of optimal end-effector size for container diameter}
The SCU-Hand was attached with a 0.2~mm transparent PP sheet, and its diameter was reconfigured to 90~mm, 80~mm, and 70~mm. All four types of containers and 10~g of flour were used. For each size container, all three sizes of end-effectors were used except the 67~mm container because the 90~mm conical sheet could not be inserted.

\subsubsection{Comparison with other end-effector}
A metal sheet and a ready-made silicon ladle were used for the end-effector; the metal sheet was less deformable than PP, and the ladle was made of silicon, so the edges of the ladle were deformed by a few millimeters upon contact. Still, no large deformation of the small container could occur. The containers with size 83~mm and 67~mm were used. The optimal size of the conical sheet was used for each container, as determined in Experiment 1. Flour was used as the scoped material.

\subsubsection{Scooping performance with other granular media}
A PP sheet was used for the end-effector, and the container size was 110~mm, with an optimum diameter of 90~mm. Coffee powder (10~g) and rice (10~g) were used as materials.

\subsection{Result}
\subsubsection{Verification of optimal end-effector size for container diameter}
Table~\ref{tb: result of deformation} shows the percentage of successfully scooped powder to the amount of powder initially placed in the container. In each cell, the upper value is the mean, and the lower indicates the standard deviation. Results in which 95\% or more were successfully scooped are colored red. 
The table shows that more than 95\% of the powder can be scoped depending on the configured size of the SCU-Hand. 
Therefore, we can conclude that each container has an optimal end-effector size.
The 90~mm end-effector had the highest scooping amount for 110~mm and 93~mm containers, and the 80~mm and 71~mm end-effectors had the highest scooping amount for containers with diameters of 80~mm and 67~mm, respectively. In actual operation during the experiment, when the end-effector is inserted into a container with a smaller diameter, although it can enter the container by deforming, it cannot completely contact the inner surface of the container, thus failing to scope up all the powder. Similarly, when the end-effector is smaller than the container, the powder spreads horizontally and spills over the sides of the end-effector.

\begin{table}[t]
  \caption{Scooping Performance Results with Varying End-Effector and Container Sizes}
  \label{tb: result of deformation}
    \centering
          \begin{minipage}{70mm}
      \centering
      \includegraphics[width=\linewidth]{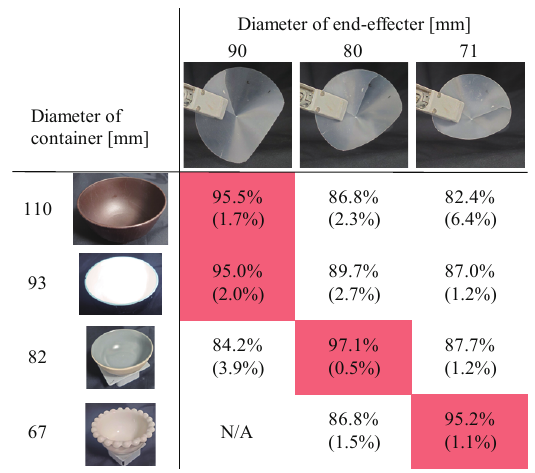}
    \end{minipage} 
\end{table}

\begin{table}[t]
  \caption{Comparison between existing tools and rigid end-effectors}
  \label{tb: result of other tool}
      \centering
          \begin{minipage}{70mm}
      \centering
      \includegraphics[width=\linewidth]{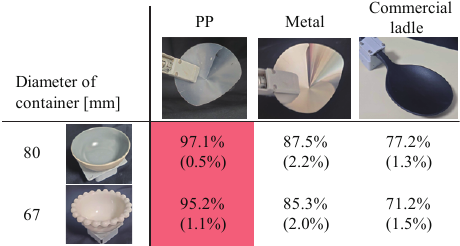}
    \end{minipage}
\end{table}

  \begin{table}[t]
  \caption{Result with granular media with different particle sizes}
  \label{tb: result of other material}
  \centering
            \begin{minipage}{70mm}
      \centering
      \includegraphics[width=\linewidth]{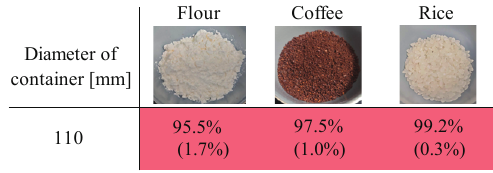}
       \end{minipage}
\end{table}

\subsubsection{Comparison with other end-effector}
Comparison with other end-effectors showed that the tool's flexibility and the resizing improved scooping performance, as shown in the results in Table~\ref{tb: result of other tool}. The results for PP are those of Experiment 1. The same shape but with a rigid metal plate and a commercially available ladle whose contact surface is flexible only resulted in a 10\% - 20\% lower scooping amount than the PP sheet version. In the actual operation, the area that does not spread horizontally and is not in contact with the container surface increases, resulting in the scrapping of the object. 

\subsubsection{Scooping performance with other granular media}
As shown in Table~\ref{tb: result of other material}, a scooping performance of more than 95\% was confirmed for all materials. In particular, the scooping performance of Rice, which has a large grain size, was close to 100\%.


\section{Discussion}
Experimental results found that the system's reconfigurability and flexibility were effective in scooping up samples from containers of different sizes. The SCU-Hand showed a scooping performance of over 95\%  in a single scooping operation for various-size containers. 
The size of the end-effector can be reduced, but as the vertex angle $\phi$ becomes sharper, contact between the frame and the container may occur, especially with deeper containers. Practically, $\phi = 90^{\circ}$, as used in this experiment, is considered optimal for ease of operation.
When the scooping trajectory is not ideal, insufficient deformation upon pressing can lead to powder spillage. While the SCU-Hand has a tolerance that does not require precise contact control, it must remain within a certain range of error relative to the optimal trajectory to prevent spillage.
As shown in Table~\ref{tb: result of deformation}, the SCU-Hand deforms asymmetrically because it uses a non-perfectly circular sheet. This may cause the diameter to be smaller than predicted by Eq.~\ref{eq_diameter}, and the reduced surface area increases the risk of powder spilling. A future challenge will be to explore the optimal sheet shape to maximize the scoop amount for a small container.

\section{Conclusions}
This study proposes a novel, flexible, and reconfigurable end-effector called SCU-Hand for powder scooping tasks. The performance of the proposed end-effector is validated through various experiments.
Comparisons with existing tools and rigid materials revealed that the proposed mechanism's flexibility contributes to performance.
The realization of such a general-purpose mechanism is expected to contribute to the automation of niche tasks that require a variety of functions.

In future work, we will pursue the introduction of measurement systems, such as vision sensors and control mechanisms, to ensure stable operation in laboratory automation tasks. 
Measuring the powder distribution and container shape could enable repeated scooping actions, potentially allowing for more efficient powder collection under a wider range of conditions.
The SCU-hand's simple design allows the flexible sheet to be replaced with various materials. For instance, for high-viscosity substances like honey or adhesives that require significant shear force, a thin and rigid material, such as a metal plate thinner than the one used in the experiment, is more suitable. 
Additionally, if the sheet is damaged, it can be easily and cost-effectively repaired by replacing the plate, allowing for long-term operation.



\section*{Acknowledgement}
We would like to thank Kanta Ono and Yusaku Nakajima for their useful discussions on the tools for material sciences. 


\bibliographystyle{IEEEtran}
\bibliography{reference}

\end{document}